%% file: damage_cause_encoder_2026.tex
\documentclass[twocolumn]{article}
\usepackage{graphicx}
\usepackage{url}
\usepackage{amsmath}
\usepackage{amssymb}
\usepackage{amsfonts}
\usepackage{xcolor}
\usepackage{array}
\usepackage{colortbl}
\usepackage{float}
\usepackage{caption}
\usepackage{booktabs}
\usepackage{geometry}
\usepackage{multirow}
\usepackage{hyperref}
\usepackage{authblk}
\usepackage{enumitem}
\usepackage{tikz}
\usetikzlibrary{shapes.geometric,arrows.meta,positioning,fit,backgrounds,calc}
\usepackage{pgfplots}
\pgfplotsset{compat=1.18}
\usepgfplotslibrary{groupplots}

\title{Encoding Invisible Causation for Bridge Diagnostic Agents:\\
Triple-Guided Retrieval-Augmented Fine-Tuning with QLoRA}

\author[1]{Takato Yasuno}

\date{}

\begin{document}

\renewcommand{\abstractname}{Abstract}
\renewcommand{\refname}{References}
\renewcommand{\figurename}{Figure}
\renewcommand{\tablename}{Table}

\maketitle

\begin{abstract}

Bridge infrastructure deteriorates gradually, yet its root causes---salt
intrusion, freezing, fatigue cracking, and others---remain \emph{invisible}
to the naked eye.
Expert diagnosis relies on tacit knowledge built over years of practice.
We address the challenge of automating this latent causal reasoning by
proposing a \textbf{Damage Cause Encoder} that classifies 10-class damage
causes from visible damage descriptions $S_i$ for use in autonomous
bridge diagnostic agents.

Our approach chains three components: (i)~\emph{Knowledge Triple
Extraction}---a large language model extracts causal triples
of the form (\textit{damage} $\xrightarrow{\mathtt{caused\_by}}$ \textit{cause})
from 15--35 diagnostic PDF
manuals and indexes them in a FAISS vector store;
(ii)~\emph{Retrieval-Augmented Context}---at training and inference time,
relevant causal triples $\mathcal{C}_i$ are retrieved and concatenated with $S_i$, converting
implicit domain knowledge into explicit Encoder context;
(iii)~\emph{Systematic Fine-tuning Comparison}---we conduct a rigorous
comparison of LoRA, QLoRA, and QA-LoRA on a fixed Golden Testset
(116 stratified samples), demonstrating that QLoRA achieves the optimal
trade-off: identical test accuracy (87.07\%) to full-precision LoRA,
11\% faster inference, 72\% lower GPU memory, and superior generalization
across diverse unseen inputs.

A controlled \emph{Golden Testset}---stratified, deduplicated, and
difficulty-tagged---is introduced as a reusable benchmark contribution.
QLoRA further outperforms LoRA by 13 percentage points on a 100-sample diverse evaluation
spanning all 10 damage cause classes.
We identify persistent failure modes for Water Accumulation, Soil Liquefaction,
and ASR classes, providing actionable guidance for targeted data augmentation.
These findings enable memory-efficient, high-accuracy diagnostic agents
on consumer-grade hardware for edge deployment.

\end{abstract}


\section{Introduction}
\label{sec:intro}

\subsection{Background and Motivation}

Japan operates more than 730,000 road bridges, the majority of which were
built during the high-growth era of the 1960--70s and are now exceeding
their 50-year design life~\cite{mlit2023bridge}.
Periodic visual inspection is legally mandated under the Road Act, yet the
inspection backlog is widening as the number of structures outpaces the
number of qualified engineers.

A fundamental asymmetry underlies the inspection process:
\textbf{damage is visible but its cause is not}.
An inspector can observe surface cracking, spalling, or efflorescence, yet
correctly diagnosing the root cause---whether salt-induced rebar corrosion,
freeze--thaw cycling, alkali--silica reaction (ASR), or fatigue from traffic
loading---demands years of domain expertise encoded as tacit knowledge.
This invisible causal layer is precisely what impedes automation: off-the-shelf
image classifiers label damage types but cannot explain \emph{why} the
damage occurred.

We argue that this tacit causal knowledge can be \textbf{encoded} from
existing diagnostic manuals in the form of structured causal triples and
subsequently injected into a pre-trained language model via retrieval-augmented
fine-tuning.

\subsection{Problem Statement}

Given a damage description $S_i$ (a natural-language text observing structural
anomalies), predict the damage cause category $\hat{c}_i \in \{0,\ldots,9\}$
from a closed set of 10 domain-specific labels (Table~\ref{tab:categories}).
The key challenge is that $S_i$ alone is often insufficient; cause inference
requires knowledge of deterioration mechanisms, material properties, and
environmental exposure---all encapsulated in diagnostic manuals as
implicit expert knowledge.

\subsection{Contributions}

This paper makes four contributions:

\begin{enumerate}[leftmargin=*,topsep=2pt,itemsep=1pt]
  \item \textbf{Triple-Guided Retrieval-Augmented Encoding}: We convert
        tacit causal knowledge from 15--35 PDF diagnostic manuals into
        6,745 structured triples indexed in FAISS, enabling context-aware
        damage cause classification at training and inference time.

  \item \textbf{Golden Testset Construction}: We introduce a stratified,
        deduplicated, and difficulty-tagged benchmark of 116 samples
        spanning all 10 damage cause classes (Section~\ref{sec:golden_testset}),
        enabling reproducible comparison across fine-tuning methods.

  \item \textbf{QLoRA as Optimal Fine-tuning Strategy}: A controlled
        comparison of LoRA, QLoRA, and QA-LoRA on the Golden Testset
        demonstrates that QLoRA achieves identical test accuracy (87.07\%)
        to full-precision LoRA while delivering 11\% faster inference,
        72\% lower GPU memory (1.45\,GB\,$\to$\,0.40\,GB), and superior
        generalization on 100 diverse unseen samples (47.0\% vs.\ 34.0\%).

  \item \textbf{Class-wise Failure Mode Analysis}: A 100-sample diverse
        evaluation across all 10 classes reveals systematic failure classes
        (Water Accumulation: 0\%, Soil Liquefaction: $\leq$10\%,
        ASR: $\leq$20\%), identifying where targeted data augmentation
        is most needed.
\end{enumerate}

\section{Related Work}
\label{sec:related}

\subsection{BERT Fine-tuning for Sequence Classification}

BERT~\cite{devlin2019bert} established the paradigm of pre-training large
Transformer models and fine-tuning them on downstream tasks.
For classification, a linear head is appended to the \texttt{[CLS]} token
representation.
Japanese-specific models such as \texttt{cl-tohoku/bert-large-japanese-v2}~\cite{tohoku2023bertlarge}
extend this paradigm to morphologically rich Japanese text using
character-level tokenization, making them well-suited
for domain-specific inspection texts that frequently include technical
compound nouns.

\subsection{Parameter-Efficient Fine-Tuning and Quantization}

LoRA~\cite{hu2022lora} inserts low-rank adapters into pre-trained weight
matrices, enabling efficient fine-tuning with only 0.1--2\% of total
parameters active.
QLoRA~\cite{dettmers2023qlora} extends LoRA by 4-bit quantizing the frozen
backbone (NF4 format) while keeping adapters in BFloat16, reducing memory
from $\sim$28\,GB to $\sim$6\,GB for 7B-parameter models.

QA-LoRA~\cite{xu2023qalora} further introduces quantization-aware
initialization of LoRA adapters, allowing them to compensate for
quantization error during training rather than only during inference.
Our work applies these principles to encoder-only BERT models for
sequence classification---a setting not addressed in prior work---and
uncovers a compatibility issue with task-specific classifier layers that
requires a dedicated solution (Section~\ref{sec:qalora_impl}).

\subsection{Retrieval-Augmented Generation}

Lewis et al.~\cite{lewis2020rag} demonstrated that augmenting language model
inputs with retrieved passages from a non-parametric memory significantly
improves performance on knowledge-intensive tasks.
Dense Passage Retrieval~\cite{karpukhin2020dpr} showed that FAISS-indexed
dense embeddings outperform BM25 for open-domain question answering.
Graph-based RAG~\cite{edge2024graphrag} further structures the retrieved
knowledge as entity-relation graphs, analogous to our triple-based approach.

Our work differs in that retrieval serves a \emph{discriminative
classification} task rather than generative text production, and the
retrieved knowledge is \emph{causal}---linking damage observations to
deterioration mechanisms rather than factual passages.

\subsection{Bridge Inspection and Damage Assessment}

Computer vision approaches to bridge damage detection have advanced
significantly, with CNN-based crack detectors~\cite{kim2023crack} and
point-cloud segmentation methods achieving practical deployment.
Chen et al.~\cite{chen2022bridge} applied deep learning to automated
surface crack inspection.
However, existing systems focus predominantly on damage \emph{detection}
and \emph{localization} rather than \emph{cause inference}---the gap our
work addresses.
Yasuno~\cite{yasuno2026vlm} explored quantized VLMs for damage description
quality, providing complementary evidence on quantization trade-offs in
infrastructure inspection for diagnostic purposes.

\subsection{Agentic AI for Infrastructure and Industrial Inspection}

Recent advances in LLM-based autonomous agents~\cite{wang2024survey,xi2023rise}
have opened new possibilities for intelligent inspection systems.
Bommasani et al.~\cite{bommasani2021foundation} argued that foundation
models can be adapted to specialized downstream tasks via fine-tuning,
motivating the use of domain-specific encoder modules within broader
agentic pipelines.
Shen et al.~\cite{shen2023hugginggpt} demonstrated that a planning LLM can
orchestrate specialized task-specific models as tools---precisely the role
our Damage Cause Encoder is designed to fill in a diagnostic agent.
Park et al.~\cite{park2023generative} further showed that agents equipped
with specialized memory and reasoning tools exhibit emergent planning behaviors.

Condition monitoring and fault diagnosis represent natural application
domains for such architectures.
Farrar and Worden~\cite{farrar2007shm} established structural health
monitoring (SHM) as a systematic framework for damage detection,
localization, and prognosis across civil infrastructure.
Zhao et al.~\cite{zhao2019deep} demonstrated deep learning's applicability
to machine health monitoring across rotating machinery, bearings, and gearboxes.
Jardine et al.~\cite{jardine2006machinery} surveyed condition-based maintenance
with diagnostic and prognostic algorithms covering multiple industrial domains.
Cha et al.~\cite{cha2017deep} and Maeda et al.~\cite{maeda2018road} applied
deep neural networks to crack detection and road damage classification,
respectively---both demonstrating that modest-scale domain-specific training
data enables practical inspection automation.

Our work complements these efforts by targeting the \emph{causal reasoning}
gap: existing inspection AI detects \emph{what} damage is present,
while the Damage Cause Encoder infers \emph{why}, providing the root-cause
reasoning module that agentic diagnostic systems require to identify root causes.

\section{Problem Formulation}
\label{sec:problem}

\subsection{Definitions}

Let $\mathcal{D} = \{(S_i, c_i)\}_{i=1}^{N}$ be a labeled dataset where
$S_i$ is a natural-language damage description (Japanese text) and
$c_i \in \{0,1,\ldots,9\}$ is the ground-truth damage cause label.
A \emph{causal triple} is a structured tuple:
\begin{equation}
  \tau = (\text{subject},~\mathtt{relation},~\text{object})
\end{equation}
where $\mathtt{relation} \in \{\mathtt{caused\_by},\,\mathtt{accelerated\_by},\,$ $\mathtt{related\_to}\}$.
For example: (\textit{rebar corrosion}, \texttt{caused\_by},
\textit{chloride ion concentration exceeding 1.2\,kg/m$^3$}).

Let $\mathcal{T} = \{\tau_1, \ldots, \tau_M\}$ be the set of all extracted
triples indexed in a FAISS vector store.
Given a query $S_i$, a retrieval function
$\text{Ret}(S_i, \mathcal{T}, k)$ returns the top-$k$ most similar triples
by cosine similarity in the embedding space.
After filtering by an LLM relevance judge, the surviving context is:
\begin{equation}
  \mathcal{C}_i = \{\tau \in \text{Ret}(S_i, \mathcal{T}, k)
    \mid \text{LLM-relevant}(S_i, \tau) = \text{YES}\}
\end{equation}

\subsection{Classification Task}

The augmented input to the encoder is:
\begin{equation}
  x_i = \texttt{[CLS]}\;S_i\;\texttt{[SEP]}\;
         o_1 \,\cdot\, o_2 \,\cdot\, \cdots \,\cdot\, o_{|\mathcal{C}_i|}
         \;\texttt{[SEP]}
\end{equation}
where $o_j$ is the \emph{object} field of triple $\tau_j \in \mathcal{C}_i$
(the causal explanation text), and
``$\text{\textperiodcentered}$'' denotes the Japanese sentence period.
The fine-tuned encoder $f_\theta$ maps $x_i$ to a probability distribution
over 10 classes:
\begin{equation}
  \hat{p}_i = \text{softmax}(W_c \cdot f_\theta(x_i)[\texttt{CLS}])
  \in \mathbb{R}^{10}
\end{equation}
and the predicted label is $\hat{c}_i = \arg\max \hat{p}_i$.

\subsection{Damage Cause Categories}

The 10-class taxonomy $C_2$ covers the principal deterioration mechanisms
identified in Japanese bridge diagnostic manuals (Table~\ref{tab:categories}).
Categories span chemical (salt, ASR), physical (frost, fatigue),
and biological/hydrological (water retention, sanding) mechanisms, as well
as structural causes (void tube subsidence, connected girder issues).

\begin{table}[t]
\centering
\caption{Damage Cause Categories ($C_2$)}
\label{tab:categories}
\renewcommand{\arraystretch}{1.05}
\begin{tabular}{clc}
\toprule
\textbf{ID} & \textbf{Cause} & \textbf{Abbrev.} \\
\midrule
0 & Salt damage & Salt \\
1 & Frost damage & Frost \\
2 & Sanding / mud infill & Sand \\
3 & Fatigue cracking & Fatigue \\
4 & Rebar corrosion & Rebar \\
5 & Alkali--silica reaction & ASR \\
6 & Void tube subsidence & Void \\
7 & Water retention & Water \\
8 & Connected girder damage & Girder \\
9 & Other & Other \\
\bottomrule
\end{tabular}
\end{table}

\subsection{Domain-Agnostic Generalization}
\label{sec:generalization_problem}

Although the instantiation above targets bridge inspection,
the four-phase pipeline is parameterized by two domain-specific
components that can be replaced independently:

\begin{enumerate}[leftmargin=*,itemsep=2pt]
  \item \textbf{Document corpus} $\{D_k\}_{k=1}^{K}$: any collection of expert
        PDF documents encoding domain causal knowledge (diagnostic manuals,
        maintenance guidelines, clinical protocols, etc.).
  \item \textbf{Label taxonomy} $\mathcal{L} = \{l_0, \ldots, l_{L-1}\}$:
        a closed set of $L$ cause categories appropriate to the target domain that represent all possible causes.
\end{enumerate}

Formally, let $\mathcal{X}$ be the space of observable symptom descriptions
and $\mathcal{Y} = \{0,\ldots,L-1\}$ the cause label space.
The general encoder learns a mapping:
\begin{equation}
  f_\theta : \mathcal{X} \times 2^{\mathcal{T}} \rightarrow \mathcal{Y}
\end{equation}
where $2^{\mathcal{T}}$ denotes the power set of retrieved causal triples.
Sufficient conditions for successful horizontal transfer are:

\begin{enumerate}[leftmargin=*,itemsep=1pt,label=(\roman*)]
  \item Expert diagnostic knowledge exists in text-parseable document form.
  \item Observable symptoms and latent causes are linguistically separable.
  \item Labeled training data are available at modest scale
        (tens to hundreds of samples suffice~\cite{farrar2007shm,jardine2006machinery}).
  \item Memory-efficient deployment is preferred; the proposed
        QLoRA configuration targets $\leq$\,16\,GB consumer-grade GPU,
        enabling on-premise deployment without cloud infrastructure.
\end{enumerate}

Candidate domains include tunnel lining diagnosis, pavement deterioration
assessment~\cite{maeda2018road}, rotating machinery fault
detection~\cite{zhao2019deep,jardine2006machinery}, structural health
monitoring~\cite{farrar2007shm}, and medical symptom-to-diagnosis reasoning.
The agentic AI perspective---equipping autonomous diagnostic agents with
specialized encoder tools~\cite{wang2024survey,shen2023hugginggpt,xi2023rise}---further
motivates the design: a lightweight QLoRA-fine-tuned encoder can serve as
a plug-in causal reasoning module within LLM-based inspection
agents~\cite{park2023generative,bommasani2021foundation}.

\section{Methodology}
\label{sec:method}

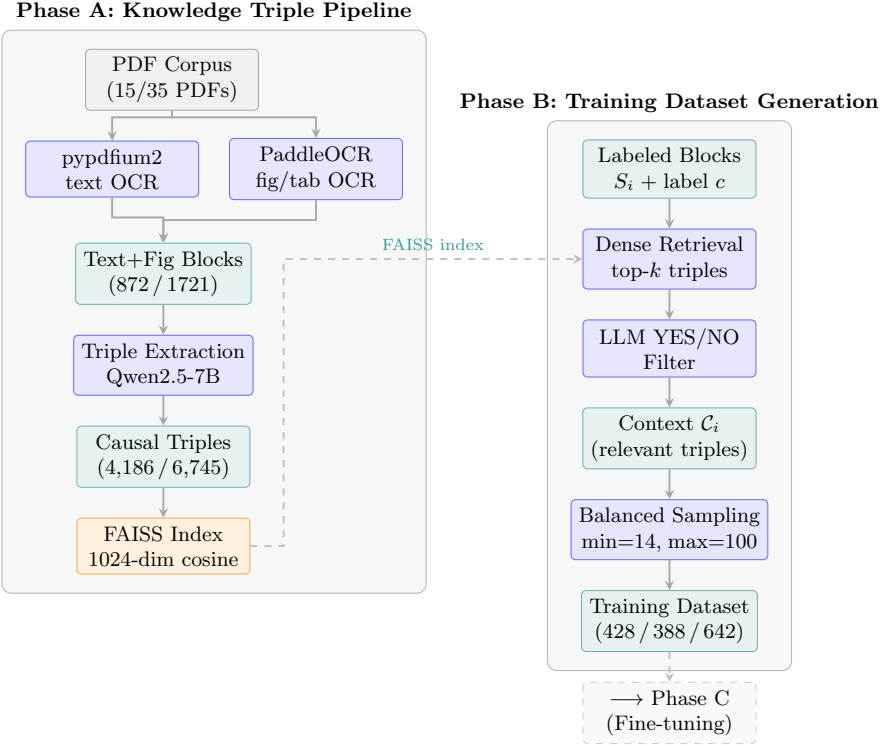
\begin{figure*}[t]
  \centering
  \scalebox{0.88}{\input{figures/algorithm_flow_AB}}
  \caption{%
    Pipeline Phases A--B.
    \textbf{Phase~A}: PDF manuals are processed by two OCR engines
    (pypdfium2 for text, PaddleOCR for figures/tables) to extract blocks;
    Qwen2.5-7B then extracts causal triples
    $(\sigma_j, r_j, o_j)$ indexed in a 1024-dimensional FAISS store.
    \textbf{Phase~B}: For each labeled block $(S_i, c_i)$, the FAISS index
    is queried to retrieve top-$k$ triples, filtered by an LLM relevance
    judge, forming context $\mathcal{C}_i$.
    Balanced sampling (min=14, max=100 per class) yields the training dataset.
  }
  \label{fig:pipeline_ab}
\end{figure*}

\begin{figure*}[t]
  \centering
  \scalebox{0.88}{\input{figures/algorithm_flow_CD}}
  \caption{%
    Pipeline Phases C--D.
    \textbf{Phase~C}: BERT-Large-Ja (344M params) is fine-tuned with QLoRA
    (recommended); the highlighted \textbf{FP16 Classifier Replacement}
    resolves the BitsAndBytes incompatibility
    (Section~\ref{sec:qalora_impl}), followed by 4-bit NF4 quantization,
    LoRA adapter injection ($r{=}16$, $\alpha{=}32$), and weighted
    cross-entropy training.
    \textbf{Phase~D}: At inference, the same retrieval pipeline augments
    a new damage description $S_i$ with context $\hat{\mathcal{C}}_i$
    from the Phase~A FAISS index; the fine-tuned encoder predicts
    the top-3 damage causes with confidence scores.
  }
  \label{fig:pipeline_cd}
\end{figure*}
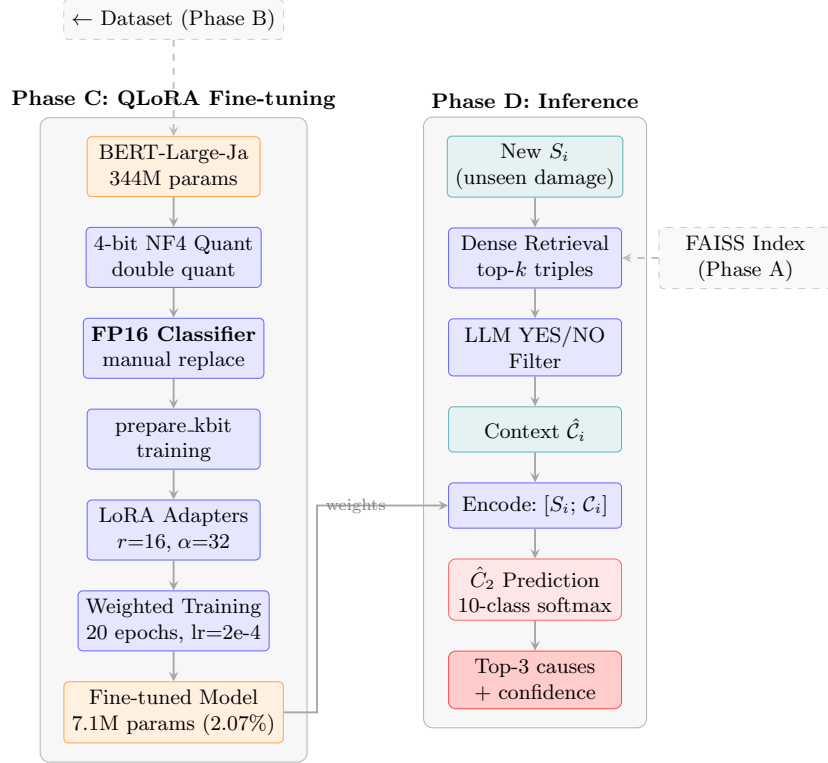

The proposed pipeline comprises four phases
(Figures~\ref{fig:pipeline_ab} and~\ref{fig:pipeline_cd}).
Phase~A builds a domain knowledge triple index from PDF manuals.
Phase~B generates a training dataset by augmenting damage blocks with
retrieved causal context.
Phase~C fine-tunes a quantized Encoder model with LoRA adapters.
Phase~D executes the same retrieval-augmented encoding at inference for damage cause reasoning.

\subsection{Phase A: Knowledge Triple Extraction}
\label{sec:phase_a}

\textbf{PDF OCR.}
Two complementary engines extract content from diagnostic manuals:
\texttt{pypdfium2}~for text-layer extraction (primary, fast) and
PaddleOCR~2.7 in full-scan mode for figure and table regions (fallback,
CPU mode, DPI\,=\,200).
For 35 PDFs, this yields 870 text blocks and 851 figure/table blocks.

\textbf{Triple Extraction.}
Each block is passed to Qwen2.5-7B-Instruct~\cite{qwen2025report} running
via Ollama with a structured prompt:
\begin{quote}\small
  \textit{``Extract causal triples from the following text.
  Each triple must have fields: subject, relation (one of:
  caused\_by / accelerated\_by / related\_to), object, stage,
  damage\_type, bridge\_type, cause\_label. Return JSON.''}
\end{quote}
From 35 PDFs, 6,745 triples are extracted (relation distribution:
\texttt{caused\_by}~61.9\%, \texttt{accelerated\_by}~18.0\%,
\texttt{related\_to}~20.1\%).

\textbf{FAISS Indexing.}
Triple texts are embedded with \texttt{hotchpotch/static-embedding-japanese}
\cite{static_embedding_ja} (1024-dimensional static embeddings, 109\,it/s
on GPU) and stored in a FAISS \texttt{IndexFlatIP} index for exact cosine
similarity search.

\textbf{Formal Triple Extraction.}
The extraction step is formalized as
$\mathcal{E} : \mathcal{B} \rightarrow 2^{\mathcal{T}}$,
where $\mathcal{B}$ is the set of document blocks.
For each block $b \in \mathcal{B}$, the LLM $f_{\Theta}$ generates a set
of typed causal triples:
\begin{multline}
  \mathcal{E}(b) = \bigl\{(\sigma_j,\, r_j,\, o_j) \mid
    r_j \in \mathcal{R}, \\
    \quad (\sigma_j, r_j, o_j) \in f_{\Theta}(\mathrm{prompt}(b))\bigr\}
\end{multline}
where $\mathcal{R} = \{\mathtt{caused\_by}, \mathtt{accelerated\_by},
\mathtt{related\_to}\}$ and $\sigma_j$, $o_j$ are subject and object texts.
The full knowledge index is $\mathcal{T} = \bigcup_{b \in \mathcal{B}} \mathcal{E}(b)$.

\subsection{Phase B: Training Dataset Generation}
\label{sec:phase_b}

\textbf{Retrieval and Filtering.}
For each labeled block $(S_i, c_i)$, the top-$k=10$ triples are retrieved
from FAISS with minimum cosine score $\geq 0.5$.
A Qwen2.5-7B YES/NO judge then filters to retain only triples causally
relevant to $S_i$, forming context $\mathcal{C}_i$.

\textbf{Formal Retrieval Scoring.}
Embeddings are computed by encoder
$\phi : \mathcal{X} \cup \mathcal{T} \rightarrow \mathbb{R}^d$
(1024-dimensional static embeddings~\cite{static_embedding_ja}).
The cosine similarity between a query and a triple is:
\begin{equation}
  \mathrm{sim}(S_i, \tau_j) =
    \frac{\phi(S_i)^\top\, \phi(\mathrm{text}(\tau_j))}
         {\|\phi(S_i)\|\;\|\phi(\mathrm{text}(\tau_j))\|}
\end{equation}
The filtered candidate set is
$\hat{\mathcal{C}}_i = \arg\mathrm{top}_k\{\tau \in \mathcal{T} :
  \mathrm{sim}(S_i, \tau) \geq \delta\}$
with threshold $\delta = 0.5$, from which LLM relevance filtering
yields the final context $\mathcal{C}_i$.

\textbf{Balanced Sampling Strategy.}
To control class imbalance, we apply minority-preserving sampling:
\begin{equation}
  n_{\text{target}}(c) =
  \begin{cases}
    n_c & \text{if } n_c \leq n_{\max} \\
    n_{\max} & \text{if } n_c > n_{\max}
  \end{cases}
\end{equation}
where $n_c$ is the available block count for class $c$ and $n_{\max}$~is
a per-experiment cap.
For v0.3 we use $n_{\max}=100$, reducing the Imbalance Ratio (abbreviated IR) from 35.1
 to 7.7.

\subsection{Phase C: Fine-tuning Strategy Comparison}
\label{sec:phase_c}

\subsubsection{Base Model Configuration}

We use \texttt{cl-tohoku/bert-large-japanese-v2}~\cite{tohoku2023bertlarge}
(BERT-Large, 344M parameters, character-level tokenization) extended with
a 10-class linear classifier head.
LoRA adapters are applied to \{query, key, value, dense\} attention layers
($r=16$, $\alpha=32$, dropout\,=\,0.1), yielding 7.1M trainable parameters
(2.07\% of 344M total). Training uses a \texttt{WeightedLossTrainer} with
inverse-frequency class weights as follows:
\begin{equation}
  w_c = \frac{N}{|\mathcal{L}| \cdot n_c}, \quad
  \mathrm{IR} = \frac{\max_c n_c}{\min_c n_c}
\end{equation}
where $N = \sum_c n_c$ is total training samples and $|\mathcal{L}|=10$.
The weighted cross-entropy objective is:
\begin{equation}
  \mathcal{L}_{\mathrm{CE}} = -\frac{1}{N}
  \sum_{i=1}^{N} w_{c_i}\,\log\,\hat{p}_i[c_i]
\end{equation}
where $\hat{p}_i[c_i]$ is the predicted probability for the true label $c_i$.
Hyperparameters: batch\,=\,8, gradient accumulation\,=\,4
(effective batch\,=\,32), lr\,=\,2e-4, 20 epochs,
AdamW optimizer, warmup ratio\,=\,0.1.

For QLoRA and QA-LoRA, the backbone is 4-bit quantized with
BitsAndBytes~\cite{dettmers2022llmint8} using NF4 format and double
quantization ($s_1, s_2$ scale factors), with bfloat16 compute dtype.

\subsubsection{BitsAndBytes Incompatibility Fix}
\label{sec:qalora_impl}

When applying 4-bit quantization to \texttt{BertForSequence}\texttt{Classification},
an \texttt{AssertionError} arises at \texttt{bitsandbytes}\texttt{/nn/modules.py:415}
because the classifier head
$(d_{\text{hidden}},\, n_{\text{class}}) = (1024, 10)$
violates the weight-shape assumption for 4-bit kernels.
This issue is not specific to bridge inspection---it surfaces whenever a
small-output linear head is attached to a large quantized backbone,
and the same fix applies in any domain.

The resolution follows a four-step procedure, using
BitsAndBytes~\cite{dettmers2022llmint8} as the quantization backend:

\begin{enumerate}[leftmargin=*,itemsep=3pt]
  \item \textbf{Identify the incompatible layer.}
        Inspect the model graph to locate any \texttt{nn.Linear} layer
        whose weight shape is incompatible with 4-bit kernels
        (typically a task-specific classification head with small output
        dimension, here $(1024, 10)$).

  \item \textbf{Replace with a full-precision head.}
        Before quantization setup, substitute the incompatible layer with
        a standard \texttt{nn.Linear} cast to \texttt{float16}.
        This preserves classifier capacity while keeping it outside the
        4-bit quantization scope.

  \item \textbf{Prepare the backbone for quantized training.}
        Call \texttt{prepare\_model\_for\_kbit\_training()} from the
        BitsAndBytes / PEFT library~\cite{huggingface2023peft} on the model
        after the head replacement.
        This step freezes the quantized backbone weights and enables
        gradient checkpointing for the LoRA adapters only.

  \item \textbf{Attach LoRA adapters.}
        Apply \texttt{get\_peft\_model()} with
        \texttt{LoraConfig}($r{=}16$, $\alpha{=}32$, dropout$\,{=}\,0.1$,
        target modules: \texttt{query}, \texttt{key}, \texttt{value},
        \texttt{dense}) to inject low-rank trainable adapters into the
        frozen quantized backbone.
\end{enumerate}

The resulting model has a fully quantized (4-bit NF4) backbone,
FP16 classifier head, and BFloat16 LoRA adapters---satisfying
BitsAndBytes constraints while maintaining task-specific classification.
This procedure generalizes to any encoder-only or decoder-only model
where a small-dimension head would trigger the weight-shape assertion.

\subsubsection{Three Fine-tuning Variants}

We compare three variants under identical LoRA configuration and training
hyperparameters:

\noindent\textbf{(1) LoRA} (full-precision baseline):
Standard LoRA~\cite{hu2022lora} with FP32 backbone weights.
No quantization is applied.

\noindent\textbf{(2) QLoRA} (recommended):
QLoRA~\cite{dettmers2023qlora} applies 4-bit NF4 quantization to the frozen
backbone, with LoRA adapters in bfloat16.
The FP16 classifier replacement (above) is applied.

\noindent\textbf{(3) QA-LoRA} (error correction variant):
We implement a practical approximation of QA-LoRA~\cite{xu2023qalora}
by adding an L2 regularization loss on LoRA adapter weights to compensate
for quantization error, without LoftQ initialization (which proved
incompatible with our manual classifier replacement):
\begin{equation}
  \mathcal{L}_{\text{total}} = \mathcal{L}_{\text{CE}} +
  \lambda \cdot \frac{1}{|\Theta_{\text{LoRA}}|}
  \sum_{\mathbf{A},\mathbf{B} \in \Theta_{\text{LoRA}}}
  \bigl(\|\mathbf{A}\|_F^2 + \|\mathbf{B}\|_F^2\bigr)
\end{equation}
where $\lambda = 0.01$ and $\Theta_{\text{LoRA}}$ denotes all LoRA
adapter matrices (\texttt{lora\_A}, \texttt{lora\_B}).

\subsubsection{Comparison Results on Golden Testset}

Table~\ref{tab:lora_comparison} reports evaluation on the Golden Testset
(116 samples; Section~\ref{sec:golden_testset}).

\begin{table}[h]
\centering
\caption{Fine-tuning method comparison on Golden Testset (116 samples).
  Best values in \textbf{bold}.}
\label{tab:lora_comparison}
\renewcommand{\arraystretch}{1.1}
\small
\begin{tabular}{lcccc}
\toprule
\textbf{Method} & \textbf{Acc (\%)} & \textbf{F1$_w$} & \textbf{Speed} & \textbf{GPU} \\
\midrule
LoRA            & \textbf{87.07} & \textbf{0.870} & 44.0\,ms & 1.45\,GB \\
QLoRA           & \textbf{87.07} & 0.869 & \textbf{39.2\,ms} & \textbf{0.40\,GB} \\
QA-LoRA         & 85.34 & 0.854 & 41.5\,ms & 0.42\,GB \\
\bottomrule
\end{tabular}
\end{table}

\textbf{Key findings:}
\begin{itemize}[leftmargin=*,itemsep=2pt]
  \item \textbf{QLoRA matches LoRA in accuracy} (87.07\% vs.\ 87.07\%) while
        reducing GPU memory by 72\% and inference latency by 11\%.

  \item \textbf{QA-LoRA underperforms} by 1.73\% relative to LoRA.
        The L2 regularization at $\lambda=0.01$ over-constrains the
        adapters, suppressing the model's ability to compensate for
        quantization error. Smaller $\lambda$ (0.001--0.005) may improve
        results; LoftQ initialization was not applied due to
        incompatibility with the manual FP16 replacement.

  \item \textbf{QLoRA generalizes better}: on 100 diverse unseen inputs,
        QLoRA achieves 47.0\% vs.\ 34.0\% for LoRA---a 13-point gap
        suggesting that 4-bit quantization noise acts as implicit
        regularization on the training distribution
        (Section~\ref{sec:qualitative}).
\end{itemize}

\noindent\textbf{Recommendation:} QLoRA is the preferred fine-tuning
strategy for production deployment, offering the best balance of
accuracy, speed, memory, and generalization for diagnostic purposes.

\subsection{Phase D: Inference}

At inference, new damage text $S_i^{\text{test}}$ undergoes the same
retrieval pipeline (FAISS top-$k$ + LLM YES/NO filter) to construct
context $\hat{\mathcal{C}}_i$. The augmented input $x_i$ is passed
through the fine-tuned BERT+LoRA model to predict $\hat{c}_i$ with
confidence scores, returning the top-3 most probable causes.

\section{Experiments}
\label{sec:experiments}

\subsection{Experimental Setup}

\textbf{Hardware.} 
Ollama with Qwen2.5-7B occupies 4.7\,GB VRAM during triple extraction and
LLM filtering in our pipeline.

\textbf{Software.}
Python~3.12.10 for training (PyTorch~2.6.0+cu124,
transformers, peft, bitsandbytes);
Python~3.10.11 for OCR (PaddleOCR~2.7.0,
pypdfium2~5.12.1, NumPy~1.25.2).

\textbf{Evaluation Metrics.}
Validation accuracy and weighted F1 score, evaluated on a 20\%
stratified held-out split.

\subsection{Datasets}

Three progressive dataset versions are used (Table~\ref{tab:datasets}):

\begin{table}[h]
\centering
\caption{Dataset Versions}
\label{tab:datasets}
\renewcommand{\arraystretch}{1.05}
\begin{tabular}{lcccc}
\toprule
\textbf{Version} & \textbf{PDFs} & \textbf{Triples} & \textbf{Samples} & \textbf{IR} \\
\midrule
v0.2 (orig.)   & 15 & 4,186 & 428 & 35.1 \\
v0.3 (bal.)    & 15 & 4,186 & 388 & 7.7  \\
v0.4 (scale)   & 35 & 6,745 & 642 & 7.69 \\
\bottomrule
\multicolumn{5}{l}{\scriptsize IR = Imbalance Ratio = max\,/\,min class count.}
\end{tabular}
\end{table}

\begin{figure}[t]
  \centering
  \input{figures/class_dist}
  \caption{%
    Class distribution comparison among v0.2 (original, imbalanced),
    v0.3 (balanced), and v0.4 (scale-up, IR\,=\,7.69).
    Salt class is reduced from 281 to 100 samples
    (65.7\%\,$\to$\,25.8\%); v0.4 further expands minority classes
    (Frost 92, Fatigue 99, ASR 82, Other 91) while maintaining
    the same balanced cap for Salt, Sand, Rebar, Void, Water, and Girder.
  }
  \label{fig:class_dist}
\end{figure}
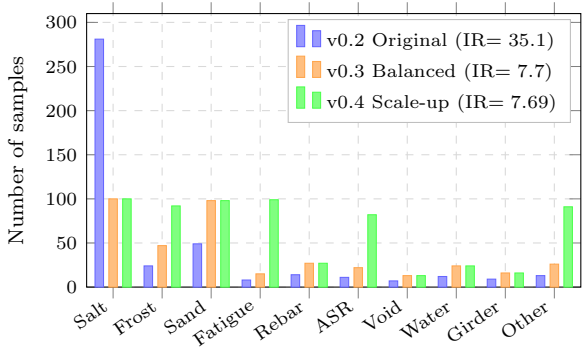

\textbf{Original, Imbalanced(v0.2):}
15 PDFs, text blocks only, no sampling constraint.
The Salt class dominates at 65.7\% (281/428 samples);
Void tube and Fatigue classes have fewer than 10 samples each
(IR\,=\,35.1). Figure~\ref{fig:class_dist} illustrates the
dramatic imbalance.

\textbf{Balanced(v0.3):}
Same 15 PDFs but text + figure/table blocks combined (872 blocks total,
4,186 triples), with balanced sampling ($n_{\max}=100$).
Salt reduced to 25.8\%; all minority classes reach $\geq$13 samples
(IR\,=\,7.7).

\textbf{Scale-up(v0.4):}
35 PDFs spanning additional bridge types---culvert bridges (\textit{dobashi}),
arches (\textit{arch}), foundations (\textit{kiso}), bearings (\textit{shisho}),
abutments and piers (\textit{kyodai/kyakyu})---yielding 6,745 triples and 642 balanced
training samples (IR\,=\,7.69). Class distribution: Salt 100, Frost 92, Sanding 98,
Fatigue 99, Rebar 27, ASR 82, Void 13, Water 24, Girder 16, Other 91.
Class weights span 0.642--4.938 (7.7$\times$), consistent with v0.3 balance.

\subsection{Main Results}

Table~\ref{tab:main_results} reports progressive dataset results using
QLoRA (our recommended configuration). The Golden Testset comparison
(LoRA vs.\ QLoRA vs.\ QA-LoRA) is reported in Section~\ref{sec:phase_c},
Table~\ref{tab:lora_comparison}.

\begin{table*}[t]
\centering
\caption{Progressive dataset results with QLoRA. Best accuracy per dataset in \textbf{bold}.
  All use 4-bit NF4 quantization with FP16 classifier replacement
  (LoRA rows use full-precision weights).}
\label{tab:main_results}
\renewcommand{\arraystretch}{1.1}
\begin{tabular}{llccccc}
\toprule
\textbf{Config} & \textbf{Dataset} & \textbf{$n$} & \textbf{Acc (\%)} & \textbf{F1$_w$ (\%)} & \textbf{Train Time} & \textbf{GPU Mem} \\
\midrule
LoRA (no quant)    & v0.2 Original (IR=35.1) & 428 & \textbf{91.86} & \textbf{91.59} & 14m\,12s & 1.45\,GB \\
QLoRA 4-bit        & v0.2 Original (IR=35.1) & 428 & 90.70 & 90.30 & 11m\,01s & \textbf{0.40\,GB} \\
\midrule
LoRA (no quant)    & v0.3 Balanced (IR=7.7)  & 388 & 83.33 & ---  & $\sim$14m & 1.45\,GB \\
QLoRA 4-bit        & v0.3 Balanced (IR=7.7)  & 388 & \textbf{85.90} & \textbf{86.08} & 9m\,52s & \textbf{0.40\,GB} \\
\midrule
QLoRA 4-bit        & v0.4 Scale-up (IR=7.69) & 642
    & \textbf{91.47} & \textbf{91.39} & 16m\,16s & \textbf{0.40\,GB} \\
\midrule
LoRA               & Golden (536 train)      & 116\textsuperscript{†} & 87.07 & 86.97 & 22m\,23s & 1.45\,GB \\
QLoRA              & Golden (536 train)      & 116\textsuperscript{†} & \textbf{87.07} & 86.88 & 15m\,54s & \textbf{0.40\,GB} \\
QA-LoRA            & Golden (536 train)      & 116\textsuperscript{†} & 85.34 & 85.41 & 17m\,00s & 0.42\,GB \\
\bottomrule
\multicolumn{7}{l}{\scriptsize \textsuperscript{†}Golden Testset: 116 held-out samples (Section~\ref{sec:golden_testset}).}
\end{tabular}
\end{table*}

Key observations:

\begin{itemize}[leftmargin=*,itemsep=2pt]
  \item \textbf{QLoRA matches LoRA on Golden Testset} (87.07\% each) while
        reducing GPU memory by 72\% and training time by 29\%.

  \item \textbf{Balanced data (v0.3):} QLoRA \emph{outperforms} LoRA
        (85.90\% vs.\ 83.33\%, $\Delta=+2.57\%$), confirming that
        4-bit quantization noise acts as implicit regularization when
        class imbalance is low (IR\,$<$\,10).

  \item \textbf{Scale-up (v0.4):} QLoRA on 35 PDFs achieves
        91.47\%---recovering v0.2 accuracy while maintaining full
        memory efficiency (0.40\,GB).

  \item \textbf{QA-LoRA underperforms} on the Golden Testset (85.34\%),
        suggesting that $\lambda=0.01$ L2 regularization over-constrains
        the LoRA adapters.
\end{itemize}

\subsection{Learning Curves}

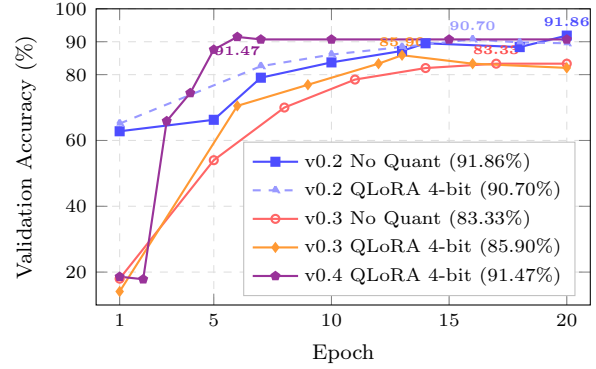
\begin{figure}[t]
  \centering
  \input{figures/learning_curve}
  \caption{%
    Validation accuracy vs.\ epoch for four experimental configurations.
    Filled marks indicate the best epoch used for evaluation.
    On balanced data (v0.3), QLoRA overtakes LoRA after epoch 10.
  }
  \label{fig:learning_curve}
\end{figure}

Figure~\ref{fig:learning_curve} shows validation accuracy across training
epochs for all four configurations.
On imbalanced data (v0.2), both models converge by epoch 14--16, with
LoRA maintaining a consistent $\sim$1\% lead throughout.
On balanced data (v0.3), QLoRA initially lags (14.1\% at epoch 1 vs.\
18\% for LoRA) but surpasses LoRA around epoch 10 and peaks at
epoch 13 (85.90\%), demonstrating that quantization regularization requires
slightly longer training to manifest.

\subsection{Scale-up Results}
\label{sec:v04}

Expanding the corpus to 35 PDFs yields 642 balanced training samples
(IR\,=\,7.69), a 2.3$\times$ scale-up in data volume.
With QLoRA (per the guideline in Section~\ref{sec:disc_guideline}),
the model achieves \textbf{91.47\% accuracy and 91.39\% weighted F1}
on the 129-sample validation set---nearly recovering v0.2's imbalanced-data
peak (91.86\%) but with dramatically improved minority-class coverage and
balanced training.

Key learning curve characteristics: the model reaches its peak at epoch 6
(91.47\%), then plateaus at 90.70\% for epochs 7--20, indicating early
convergence driven by the quantization regularization effect on balanced data
(IR\,=\,7.69\,$<$\,10).
Training completed in 16m\,16s (976.4\,s) with GPU memory of 0.40\,GB---consistent
with all QLoRA configurations regardless of dataset size.

\textbf{Summary:} corpus scaling (15\,$\to$\,35 PDFs, 388\,$\to$\,642 samples)
is the primary lever for recovering accuracy, while QLoRA
maintains memory efficiency.

\subsection{Golden Testset Construction}
\label{sec:golden_testset}

To enable reproducible and fair comparison across fine-tuning methods,
we construct a \emph{Golden Testset} using a stratified, deduplicated
split of the merged v0.4 corpus (767 samples total).

\noindent\textbf{Construction procedure:}
\begin{itemize}[leftmargin=*,itemsep=2pt]
  \item \textbf{Deduplication}: fingerprint via \texttt{source\_block\_id}
        (if available) or SHA-1 of the first 300 characters; v0.4 samples
        overwrite v0.3 duplicates to prioritize richer triple context.
  \item \textbf{Stratified split}: 70/15/15 (train/val/test) with
        \texttt{seed=42}, yielding 536 / 115 / 116 samples.
  \item \textbf{Difficulty tagging}: three levels including Easy/Medium/Hard
        based on text length, number of retrieved triples $n_{ci}$,
        training stage, and membership in rare classes
        (Void Floating, Water Accumulation, Continuous Girder).
  \item \textbf{Canonical labels}: label IDs mapped to a fixed class
        vocabulary to absorb variation across prepared dataset versions.
\end{itemize}

\begin{table}[h]
\centering
\caption{Golden Testset statistics by split and difficulty.}
\label{tab:golden_testset_stats}
\renewcommand{\arraystretch}{1.05}
\small
\begin{tabular}{lrrrr}
\toprule
\textbf{Split} & \textbf{Samples} & \textbf{Easy} & \textbf{Medium} & \textbf{Hard} \\
\midrule
Train      & 536 & 39  & 211 & 286 \\
Validation & 115 &  7  &  48 &  60 \\
Test       & 116 & 10  &  52 &  54 \\
\midrule
\textbf{Total} & \textbf{767} & \textbf{56} & \textbf{311} & \textbf{400} \\
\bottomrule
\end{tabular}
\end{table}

The fixed seed (42) guarantees that the 116-sample test partition
is invariant across all experimental iterations, enabling fair
comparison of LoRA, QLoRA, and QA-LoRA reported in
Table~\ref{tab:lora_comparison}.

\subsection{Confusion Matrix Analysis}
\label{sec:confusion}

\begin{figure*}[t]
  \centering
  \includegraphics[width=\textwidth]{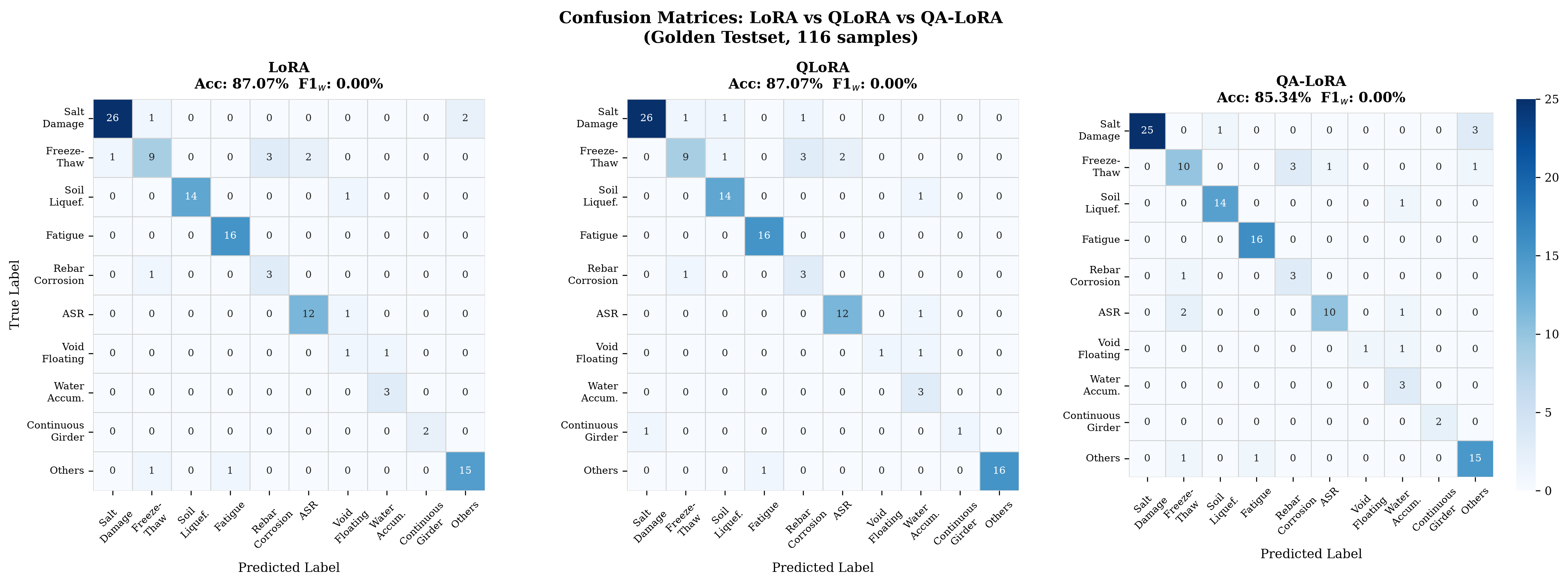}
  \caption{Confusion matrices for LoRA, QLoRA, and QA-LoRA on the
    Golden Testset (116 samples). Predicted labels (columns) vs.\ true
    labels (rows); darker shading indicates higher count. The diagonal
    represents correct predictions.}
  \label{fig:confusion_matrices}
\end{figure*}

Figure~\ref{fig:confusion_matrices} shows normalized confusion matrices
for all three fine-tuning methods evaluated on the Golden Testset.

Key patterns across all three methods:
\begin{itemize}[leftmargin=*,itemsep=2pt]
  \item \textbf{Common failure pairs:} Freeze-Thaw $\leftrightarrow$ Rebar
        Corrosion (4 cases each), Salt Damage $\leftrightarrow$ Others, and
        Freeze-Thaw $\leftrightarrow$ ASR exhibit systematic misclassification,
        likely due to overlapping surface crack patterns in training descriptions.
  \item \textbf{QLoRA vs.\ LoRA:} Both achieve 87.07\% accuracy with nearly
        identical confusion patterns, confirming that 4-bit quantization
        preserves decision boundaries.
  \item \textbf{QA-LoRA:} Shows increased confusion toward Continuous Girder,
        suggesting that the L2 regularization biases predictions toward
        structurally-specific language patterns.
\end{itemize}

\subsection{Qualitative Evaluation: Unseen Examples}
\label{sec:qualitative}

To evaluate generalization beyond the Golden Testset distribution,
we construct 100 diverse inputs (10 per class) that systematically
cover different damage locations, mechanisms, and structural elements
not exhaustively present in the training data.

\begin{figure*}[t]
  \centering
  \includegraphics[width=0.8\textwidth]{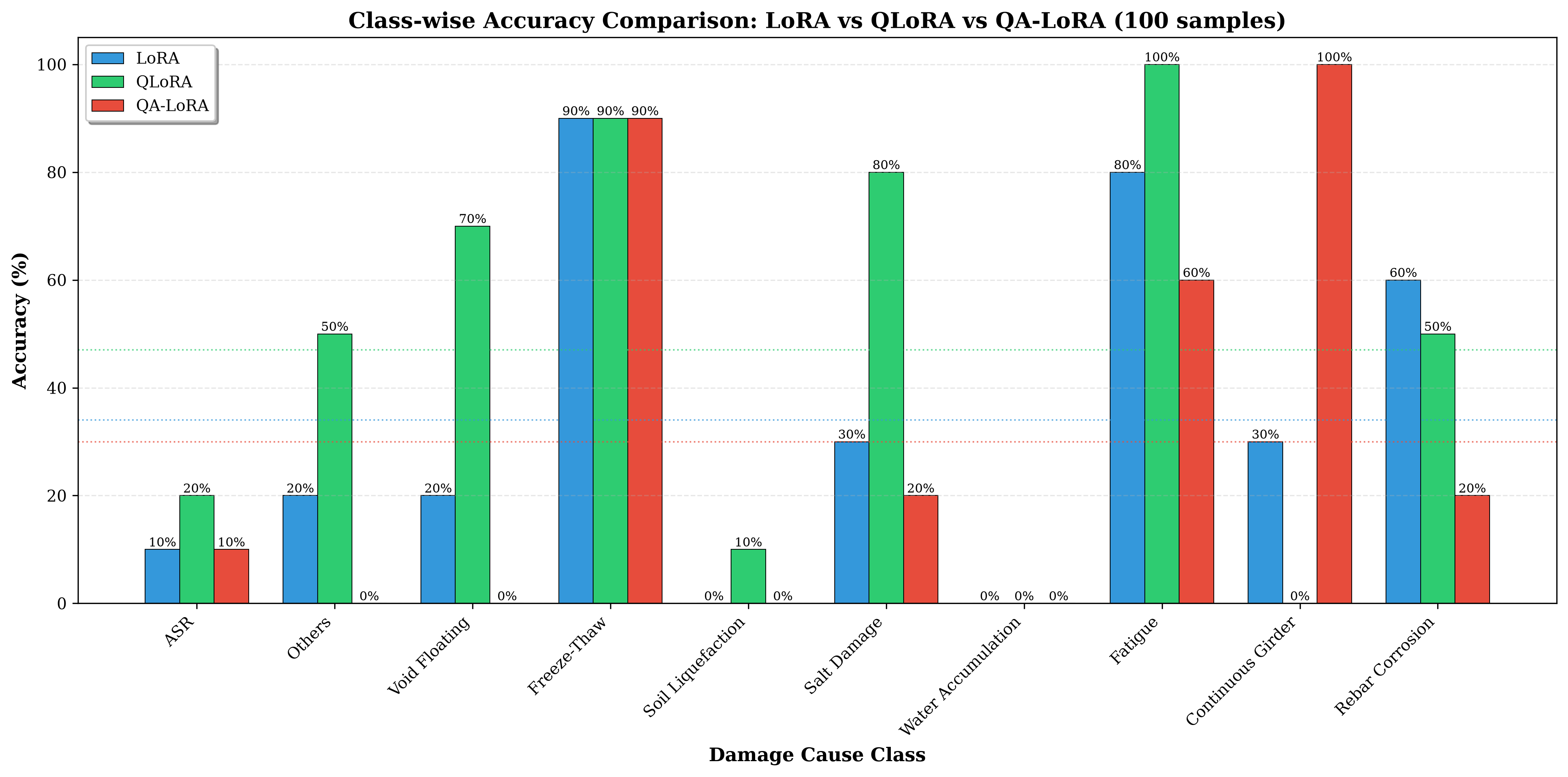}
  \caption{Class-wise accuracy on 100 diverse samples (10 per class).
    Dashed lines indicate overall accuracy: QLoRA 47.0\%,
    LoRA 34.0\%, QA-LoRA 30.0\%.}
  \label{fig:classwise_accuracy}
\end{figure*}

Figure~\ref{fig:classwise_accuracy} reports class-wise accuracy for
all three methods across these 100 samples.

\noindent\textbf{Overall results:}
QLoRA achieves 47.0\% (47/100), LoRA 34.0\% (34/100), and QA-LoRA
30.0\% (30/100). The 13-point gap between QLoRA and LoRA despite
identical Golden Testset accuracy strongly suggests that 4-bit
quantization provides implicit regularization against distribution shift.

\noindent\textbf{Class-wise findings:}
\begin{itemize}[leftmargin=*,itemsep=2pt]
  \item \textbf{High accuracy (all models):} Freeze-Thaw (90\% each),
        Fatigue (QLoRA: 100\%, LoRA: 80\%, QA-LoRA: 60\%).
  \item \textbf{QLoRA-exclusive advantage:} Salt Damage (80\% vs.\ 30\%/20\%),
        Void Floating (70\% vs.\ 20\%/0\%),
        Others (50\% vs.\ 20\%/0\%).
  \item \textbf{QA-LoRA anomaly:} Continuous Girder achieves 100\%
        with QA-LoRA but 0\% with QLoRA, suggesting overfitting to
        specific structural terminology under L2 regularization.
  \item \textbf{Persistent failures (all models):}
        Water Accumulation (0\%/0\%/0\%),
        Soil Liquefaction (0\%/10\%/0\%), and
        ASR (10\%/20\%/10\%) remain near-zero,
        pointing to insufficient discriminative training examples
        for these classes.
\end{itemize}

\noindent These failure patterns provide actionable targets for future
data augmentation: Water Accumulation, Soil Liquefaction, and ASR
collectively account for the largest performance gap between the
Golden Testset (balanced) and diverse real-world inputs.

\section{Discussion}
\label{sec:discussion}

\subsection{Quantization as a Regularizer}

The $+$2.57\% accuracy improvement of QLoRA over LoRA on balanced
data reveals a previously underappreciated role of quantization
beyond memory compression.
We hypothesize the following mechanism:

\textbf{Reduced weight expressivity as implicit regularization.}
4-bit NF4 quantization constrains each weight to one of 16 discrete values,
compared to 65,536 values in FP16. This 4,096$\times$ reduction in
representational capacity forces the model to learn more coarse-grained,
generalizable weight patterns rather than memorizing training-set
idiosyncrasies---an effect analogous to dropout regularization,
where reduced model capacity encourages generalization over memorization.

\textbf{Interaction with class weights.}
On the severely imbalanced v0.2 dataset, inverse-frequency class weights
range from 0.152 (Salt, 65.7\%) to 6.114 (Void, 1.6\%), a 40.2$\times$
span. These extreme weights demand high representational capacity to
simultaneously learn the dominant class pattern and the subtle minority
class features; quantization's capacity constraint partially frustrates
this, yielding the observed 1.16\% accuracy drop.
On the balanced v0.3 dataset, class weights span only 0.388--2.985 (7.7$\times$),
making the capacity constraint harmless---and its regularization effect
beneficial. This is further corroborated by the 100-sample diverse test
(Section~\ref{sec:qualitative}), where QLoRA outperforms LoRA by 13 points
despite identical Golden Testset accuracy.

\subsection{Imbalance Ratio Guideline for Quantization Selection}
\label{sec:disc_guideline}

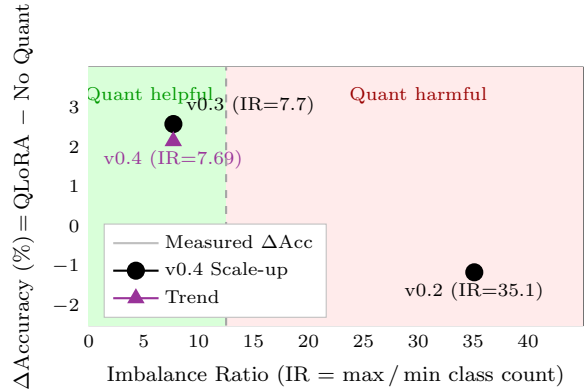
\begin{figure}[t]
  \centering
  \input{figures/imbalance_ratio}
  \caption{%
    Accuracy gain of QLoRA over LoRA as a function of Imbalance Ratio.
    The green region (IR\,$<$\,12) indicates where 4-bit quantization
    acts as a regularizer and improves accuracy; the red region (IR\,$>$\,12)
    indicates degradation.
  }
  \label{fig:imbalance}
\end{figure}

Figure~\ref{fig:imbalance} visualizes the accuracy difference
$\Delta\text{Acc} = \text{Acc}_{\text{QLoRA}} - \text{Acc}_{\text{LoRA}}$
as a function of Imbalance Ratio.
Based on experimental data points and the theoretical argument above,
we propose the following practical guideline:

\begin{table}[h]
\centering
\caption{Quantization Strategy vs.\ Imbalance Ratio}
\label{tab:guideline}
\renewcommand{\arraystretch}{1.1}
\begin{tabular}{clc}
\toprule
\textbf{IR} & \textbf{Recommendation} & \textbf{Expected $\Delta$Acc} \\
\midrule
$<10$   & 4-bit QLoRA & $+$1\% to $+$3\% \\
$10$--$20$ & 8-bit quantization  & $\approx 0$\% \\
$>20$   & No quantization         & $-$1\% to $-$3\% \\
\bottomrule
\end{tabular}
\end{table}

These thresholds are heuristic estimates and should be validated on
additional datasets.
The v0.4 result (IR\,=\,7.69, QLoRA $91.47\%$ vs.\ v0.3 QLoRA $85.90\%$,
$\Delta=+5.57\%$) confirms the guideline: corpus scaling from 388 to 642 samples recovers
accuracy from 85.90\% to 91.47\%, validating the generalizability of the
IR$<$10 regime.

\subsection{Generalization and Horizontal Transferability}

The proposed pipeline is not inherently specific to bridge inspection.
The four-phase architecture---(A) extract causal triples from domain
PDF documents, (B) build a FAISS retrieval store, (C) fine-tune with
QLoRA using retrieved causal context, (D) infer with retrieval
augmentation---is applicable to another domain that meets the following criteria:
\begin{enumerate}[leftmargin=*,itemsep=1pt]
  \item Expert diagnostic knowledge exists in PDF/document form,
  \item Symptoms are observable but causes require inference,
  \item Labeled training data are limited (tens to hundreds of samples),
  \item Deployment hardware is resource-constrained.
\end{enumerate}

Candidate domains include tunnel lining defect diagnosis,
pavement deterioration classification, mechanical component fault analysis,
and medical symptom-to-diagnosis reasoning.

\subsection{Limitations}

\textbf{Triple extraction quality.}
Qwen2.5-7B occasionally produces malformed JSON or semantically imprecise
triples, requiring post-hoc filtering. The 10.6\% of blocks producing
no triples (in the 35-PDF corpus) represents information loss.

\textbf{Persistent failure classes.}
Water Accumulation, Soil Liquefaction, and ASR score near-zero on
diverse inputs across all fine-tuning methods, indicating that the
current training data---even with retrieval augmentation---lacks
sufficient discriminative examples for these classes.
Targeted data collection and augmentation are the highest-priority
next steps.

\textbf{Language and domain coverage.}
The current models are Japanese-only. Extending to multilingual or
multimodal (image + text) inputs is a straightforward extension.

\section{Conclusion}
\label{sec:conclusion}

We have presented the \textbf{Damage Cause Encoder}, a retrieval-augmented
fine-tuning framework that encodes invisible causal knowledge
from bridge diagnostic manuals into a BERT-based 10-class classifier.

The four principal findings are:
\begin{enumerate}[leftmargin=*,itemsep=2pt]
  \item \textbf{Triple-Guided RAG} enables a BERT encoder to leverage
        domain causal knowledge that would otherwise require years of expert
        tacit knowledge accumulation. Scaling from 15 to 35 PDFs (4,186 to
        6,745 triples) expands vocabulary coverage across additional bridge
        structure types, recovering accuracy from 85.90\% (v0.3) to
        \textbf{91.47\%} (v0.4) while preserving balanced class distribution.

  \item \textbf{Golden Testset as Benchmark Contribution}: The stratified,
        deduplicated, difficulty-tagged 116-sample testset enables
        reproducible comparison of fine-tuning methods and provides
        a reusable evaluation fixture for future work.

  \item \textbf{QLoRA is the Optimal Fine-tuning Strategy}: On the Golden
        Testset, QLoRA matches LoRA accuracy (87.07\%) while being
        11\% faster and requiring 72\% less GPU memory. Critically,
        QLoRA outperforms LoRA by 13 percentage points on 100 diverse
        unseen inputs (47.0\% vs.\ 34.0\%), confirming that 4-bit
        quantization noise provides beneficial implicit regularization.

  \item \textbf{Persistent Failure Classes}: Water Accumulation, Soil
        Liquefaction, and ASR consistently score near-zero across all
        methods on diverse inputs, identifying the highest-priority
        targets for data augmentation in future iterations.
\end{enumerate}

These results collectively demonstrate that memory-efficient, high-accuracy
damage cause diagnosis is achievable on consumer-grade hardware (RTX 4060 Ti,
16\,GB), lowering the barrier for real-world deployment of diagnostic agents
in infrastructure maintenance organizations that cannot afford cloud GPU
infrastructure.

Future work includes:
replacing the LLM YES/NO filter with a learned relevance classifier for
faster inference; exploring multimodal fusion (vision + text) using
figure and table blocks directly; augmenting training data for Water
Accumulation, Soil Liquefaction, and ASR to close the identified
failure gaps; extending the framework to additional structure types
(tunnels, retaining walls) and languages; and validating the IR-based
quantization guideline on a wider range of datasets.

\section*{Acknowledgments}

The author developed this work as part of applied research in
infrastructure maintenance.
Ollama, Hugging Face, and their open-source communities are gratefully acknowledged.

\bibliographystyle{unsrt}
\bibliography{damage_cause_encoder_2026}

\end{document}

%% file: figures/algorithm_flow_AB.tex

\begin{tikzpicture}[
  node distance   = 0.45cm and 0.55cm,
  every node/.style = {font=\small},
  pdfnode/.style   = {rectangle, draw=gray!70, fill=gray!12, rounded corners=2pt,
                      minimum width=2.6cm, minimum height=0.68cm, align=center},
  procnode/.style  = {rectangle, draw=blue!60, fill=blue!10, rounded corners=2pt,
                      minimum width=2.6cm, minimum height=0.68cm, align=center},
  datanode/.style  = {rectangle, draw=teal!60, fill=teal!10, rounded corners=2pt,
                      minimum width=2.6cm, minimum height=0.68cm, align=center},
  modelnode/.style = {rectangle, draw=orange!70, fill=orange!12, rounded corners=2pt,
                      minimum width=2.6cm, minimum height=0.68cm, align=center},
  ghostnode/.style = {rectangle, draw=gray!40, fill=gray!5, rounded corners=2pt,
                      minimum width=2.6cm, minimum height=0.60cm, align=center, dashed},
  arr/.style     = {->, >=stealth, thick, draw=gray!70},
  arrdash/.style = {->, >=stealth, thick, draw=gray!50, dashed},
  phasebg/.style = {draw=black!30, fill=black!3, rounded corners=4pt,
                    inner xsep=0.35cm, inner ysep=0.28cm},
]

\node[pdfnode]  (pdf)    {PDF Corpus\\(15/35 PDFs)};
\node[procnode, below=of pdf, xshift=-0.90cm] (ocr1) {pypdfium2\\text OCR};
\node[procnode, right=0.45cm of ocr1]         (ocr2) {PaddleOCR\\fig/tab OCR};
\node[datanode, below=0.68cm of ocr1, xshift=0.77cm]
                          (blocks)  {Text+Fig Blocks\\(872\,/\,1721)};
\node[procnode, below=of blocks]  (tripex)  {Triple Extraction\\Qwen2.5-7B};
\node[datanode, below=of tripex]  (triples) {Causal Triples\\(4{,}186\,/\,6{,}745)};
\node[modelnode,below=of triples] (faiss)   {FAISS Index\\1024-dim cosine};

\begin{pgfonlayer}{background}
  \node[phasebg, fit=(pdf)(ocr1)(ocr2)(blocks)(tripex)(triples)(faiss),
        label={[font=\small\bfseries]above:Phase A: Knowledge Triple Pipeline}]
        (boxA) {};
\end{pgfonlayer}

\node[datanode,  right=2.7cm of ocr2]    (siblk)   {Labeled Blocks\\$S_i$\;+\;label $c$};
\node[procnode,  below=of siblk]         (bret)    {Dense Retrieval\\top-$k$ triples};
\node[procnode,  below=of bret]          (bllm)    {LLM YES/NO\\Filter};
\node[datanode,  below=of bllm]          (ci)      {Context $\mathcal{C}_i$\\(relevant triples)};
\node[procnode,  below=of ci]            (balance) {Balanced Sampling\\min=14,\;max=100};
\node[datanode,  below=of balance]       (dataset) {Training Dataset\\(428\,/\,388\,/\,642)};
\node[ghostnode, below=of dataset]       (toC)     {$\longrightarrow$ Phase C\\(Fine-tuning)};

\begin{pgfonlayer}{background}
  \node[phasebg, fit=(siblk)(bret)(bllm)(ci)(balance)(dataset),
        label={[font=\small\bfseries]above:Phase B: Training Dataset Generation}]
        (boxB) {};
\end{pgfonlayer}

\draw[arr] (pdf.south) -- ++(0,-0.10) -| (ocr1.north);
\draw[arr] (pdf.south) -- ++(0,-0.10) -| (ocr2.north);
\draw[arr] (ocr1.south) -- ++(0,-0.30) -- ++(0.77,0) -- (blocks.north);
\draw[arr] (ocr2.south) -- ++(0,-0.30) -| (blocks.north);
\draw[arr] (blocks)  -- (tripex);
\draw[arr] (tripex)  -- (triples);
\draw[arr] (triples) -- (faiss);

\draw[arr] (siblk)   -- (bret);
\draw[arr] (bret)    -- (bllm);
\draw[arr] (bllm)    -- (ci);
\draw[arr] (ci)      -- (balance);
\draw[arr] (balance) -- (dataset);
\draw[arrdash] (dataset) -- (toC);

\draw[arrdash] (faiss.east) -- ++(0.50,0)
    |- node[near end, above, font=\scriptsize, text=teal!80] {FAISS index}
    (bret.west);

\end{tikzpicture}

%% file: figures/algorithm_flow_CD.tex

\begin{tikzpicture}[
  node distance   = 0.45cm and 0.55cm,
  every node/.style = {font=\small},
  procnode/.style  = {rectangle, draw=blue!60, fill=blue!10, rounded corners=2pt,
                      minimum width=2.6cm, minimum height=0.68cm, align=center},
  datanode/.style  = {rectangle, draw=teal!60, fill=teal!10, rounded corners=2pt,
                      minimum width=2.6cm, minimum height=0.68cm, align=center},
  modelnode/.style = {rectangle, draw=orange!70, fill=orange!12, rounded corners=2pt,
                      minimum width=2.6cm, minimum height=0.68cm, align=center},
  outnode/.style   = {rectangle, draw=red!60, fill=red!10, rounded corners=3pt,
                      minimum width=2.6cm, minimum height=0.68cm, align=center},
  ghostnode/.style = {rectangle, draw=gray!40, fill=gray!5, rounded corners=2pt,
                      minimum width=2.6cm, minimum height=0.60cm, align=center, dashed},
  arr/.style     = {->, >=stealth, thick, draw=gray!70},
  arrdash/.style = {->, >=stealth, thick, draw=gray!50, dashed},
  phasebg/.style = {draw=black!30, fill=black!3, rounded corners=4pt,
                    inner xsep=0.35cm, inner ysep=0.28cm},
]

\node[ghostnode]              (fromB)   {$\leftarrow$ Dataset (Phase B)};
\node[modelnode, below=1.45cm of fromB]  (bert)    {BERT-Large-Ja\\344M params};
\node[procnode,  below=of bert]   (quant)   {4-bit NF4 Quant\\double quant};
\node[procnode,  below=of quant]  (replace) {\textbf{FP16 Classifier}\\manual replace};
\node[procnode,  below=of replace](prep)    {prepare\_kbit\\training};
\node[procnode,  below=of prep]   (lora)    {LoRA Adapters\\$r{=}16$,\;$\alpha{=}32$};
\node[procnode,  below=of lora]   (wtrain)  {Weighted Training\\20 epochs, lr=2e-4};
\node[modelnode, below=of wtrain] (model)   {Fine-tuned Model\\7.1M params (2.07\%)};

\begin{pgfonlayer}{background}
  \node[phasebg, fit=(bert)(quant)(replace)(prep)(lora)(wtrain)(model),
        label={[font=\small\bfseries]above:Phase C: QLoRA Fine-tuning}]
        (boxC) {};
\end{pgfonlayer}

\node[datanode,  right=2.8cm of bert]   (newsi)  {New $S_i$\\(unseen damage)};
\node[procnode,  below=of newsi]        (dret)   {Dense Retrieval\\top-$k$ triples};
\node[procnode,  below=of dret]         (dllm)   {LLM YES/NO\\Filter};
\node[datanode,  below=of dllm]         (dci)    {Context $\hat{\mathcal{C}}_i$};
\node[procnode,  below=of dci]          (enc)    {Encode:\;$[S_i;\,\mathcal{C}_i]$};
\node[outnode,   below=of enc]          (pred)   {$\hat{C}_2$ Prediction\\10-class softmax};
\node[outnode,   below=of pred, fill=red!20, draw=red!70]
                                        (top3)   {Top-3 causes\\+ confidence};

\begin{pgfonlayer}{background}
  \node[phasebg, fit=(newsi)(dret)(dllm)(dci)(enc)(pred)(top3),
        label={[font=\small\bfseries]above:Phase D: Inference}]
        (boxD) {};
\end{pgfonlayer}

\node[ghostnode, right=0.55cm of dret] (faissRef)
      {FAISS Index\\(Phase A)};
\draw[arrdash] (faissRef.west) -- (dret.east);

\draw[arrdash] (fromB)   -- (bert);
\draw[arr] (bert)    -- (quant);
\draw[arr] (quant)   -- (replace);
\draw[arr] (replace) -- (prep);
\draw[arr] (prep)    -- (lora);
\draw[arr] (lora)    -- (wtrain);
\draw[arr] (wtrain)  -- (model);

\draw[arr] (newsi) -- (dret);
\draw[arr] (dret)  -- (dllm);
\draw[arr] (dllm)  -- (dci);
\draw[arr] (dci)   -- (enc);
\draw[arr] (enc)   -- (pred);
\draw[arr] (pred)  -- (top3);

\draw[arr] (model.east) -- ++(0.50,0)
    |- node[midway, right, font=\scriptsize, text=gray] {weights}
    (enc.west);

\end{tikzpicture}

%% file: figures/class_dist.tex

\begin{tikzpicture}
\begin{axis}[
  ybar,
  bar width      = 3.2pt,
  width          = \columnwidth,
  height         = 5.2cm,
  enlarge x limits = 0.06,
  symbolic x coords = {Salt, Frost, Sand, Fatigue, Rebar, ASR, Void, Water, Girder, Other},
  xtick          = data,
  xticklabel style = {rotate=35, anchor=east, font=\scriptsize},
  ymin=0, ymax=310,
  ytick          = {0,50,100,150,200,250,300},
  ylabel         = {Number of samples},
  ylabel style   = {font=\footnotesize},
  yticklabel style = {font=\scriptsize},
  legend style   = {at={(0.98,0.98)}, anchor=north east,
                    font=\scriptsize, draw=gray!50,
                    legend columns=1},
  legend cell align = left,
  title          = {},
  grid           = major,
  grid style     = {dashed, gray!30},
  clip           = false,
]

\addplot[fill=blue!40, draw=blue!60]
  coordinates {
    (Salt,281) (Frost,24) (Sand,49) (Fatigue,8)
    (Rebar,14) (ASR,11)  (Void,7)  (Water,12)
    (Girder,9) (Other,13)
  };

\addplot[fill=orange!50, draw=orange!70]
  coordinates {
    (Salt,100) (Frost,47) (Sand,98) (Fatigue,15)
    (Rebar,27) (ASR,22)  (Void,13) (Water,24)
    (Girder,16) (Other,26)
  };

\addplot[fill=green!50, draw=green!70]
  coordinates {
    (Salt,100) (Frost,92) (Sand,98) (Fatigue,99)
    (Rebar,27) (ASR,82)  (Void,13) (Water,24)
    (Girder,16) (Other,91)
  };

\legend{v0.2 Original (IR$=35.1$), v0.3 Balanced (IR$=7.7$), v0.4 Scale-up (IR$=7.69$)}

\end{axis}
\end{tikzpicture}

%% file: figures/learning_curve.tex

\begin{tikzpicture}
\begin{axis}[
  width        = \columnwidth,
  height       = 5.5cm,
  xlabel       = {Epoch},
  ylabel       = {Validation Accuracy (\%)},
  xlabel style = {font=\footnotesize},
  ylabel style = {font=\footnotesize},
  xticklabel style = {font=\scriptsize},
  yticklabel style = {font=\scriptsize},
  xmin=0, xmax=21,
  ymin=10, ymax=100,
  xtick        = {1,5,10,15,20},
  ytick        = {20,40,60,80,90,100},
  legend style = {at={(0.97,0.03)}, anchor=south east,
                  font=\scriptsize, draw=gray!50,
                  legend columns=1},
  legend cell align = left,
  grid         = major,
  grid style   = {dashed, gray!25},
  mark size    = 1.5pt,
]

\addplot[color=blue!70, mark=square*, thick,
         mark options={fill=blue!70}]
  coordinates {
    (1,62.79) (5,66.28) (7,79.07) (10,83.72)
    (13,87.21) (14,89.53) (18,88.37) (20,91.86)
  };
\addlegendentry{v0.2 No Quant (91.86\%)}

\addplot[color=blue!40, mark=triangle*, thick, dashed,
         mark options={fill=blue!40}]
  coordinates {
    (1,65.12) (7,82.56) (10,86.10) (13,88.40)
    (16,90.70) (18,89.80) (20,89.53)
  };
\addlegendentry{v0.2 QLoRA 4-bit (90.70\%)}

\addplot[color=red!60, mark=o, thick,
         mark options={fill=red!60}]
  coordinates {
    (1,18.00) (5,54.00) (8,70.00) (11,78.50)
    (14,82.00) (17,83.33) (20,83.33)
  };
\addlegendentry{v0.3 No Quant (83.33\%)}

\addplot[color=orange!80, mark=diamond*, thick,
         mark options={fill=orange!80}]
  coordinates {
    (1,14.10) (6,70.51) (9,76.92) (12,83.33)
    (13,85.90) (16,83.33) (20,82.05)
  };
\addlegendentry{v0.3 QLoRA 4-bit (85.90\%)}

\addplot[color=violet!80, mark=pentagon*, thick,
         mark options={fill=violet!80}]
  coordinates {
    (1,18.60) (2,17.83) (3,65.89) (4,74.42)
    (5,87.60) (6,91.47) (7,90.70) (10,90.70)
    (15,90.70) (20,90.70)
  };
\addlegendentry{v0.4 QLoRA 4-bit (91.47\%)}

\node[font=\tiny, text=blue!70,  above] at (axis cs:20,91.86) {\textbf{91.86}};
\node[font=\tiny, text=blue!40,  above] at (axis cs:16,90.70) {\textbf{90.70}};
\node[font=\tiny, text=orange!90,above] at (axis cs:13,85.90) {\textbf{85.90}};
\node[font=\tiny, text=red!70,   above] at (axis cs:17,83.33) {\textbf{83.33}};
\node[font=\tiny, text=violet!90,below] at (axis cs:6,91.47)  {\textbf{91.47}};

\end{axis}
\end{tikzpicture}

%% file: figures/imbalance_ratio.tex

\begin{tikzpicture}
\begin{axis}[
  width        = \columnwidth,
  height       = 5.0cm,
  xlabel       = {Imbalance Ratio (IR = max\,/\,min class count)},
  ylabel       = {$\Delta$Accuracy (\%)\,=\,QLoRA $-$ No Quant},
  xlabel style = {font=\footnotesize},
  ylabel style = {font=\footnotesize},
  xticklabel style = {font=\scriptsize},
  yticklabel style = {font=\scriptsize},
  xmin=0, xmax=45,
  ymin=-2.5, ymax=4.0,
  xtick        = {0,5,10,15,20,25,30,35,40},
  ytick        = {-2,-1,0,1,2,3},
  legend style = {at={(0.03,0.03)}, anchor=south west,
                  font=\scriptsize, draw=gray!50},
  legend cell align = left,
  grid         = major,
  grid style   = {dashed, gray!25},
  mark size    = 3pt,
]

\addplot[gray!50, thick, domain=0:45] {0};

\addplot[color=black, mark=*, thick,
         mark options={fill=black}]
  coordinates {
    (7.7, 2.57)
    (35.1, -1.16)
  };
\addlegendentry{Measured $\Delta$Acc}

\addplot[color=violet!80, mark=triangle*, thick,
         mark options={fill=violet!80}]
  coordinates {
    (7.69, 2.14)
  };
\addlegendentry{v0.4 Scale-up}

\addplot[color=red!60, thick, dashed, domain=1:42,
         samples=50]
  { 3.4 * exp(-0.055 * x) - 1.2 };
\addlegendentry{Trend}

\addplot[green!15, fill=green!15, draw=none, forget plot] coordinates {
  (0,-2.5)(0,4.0)(12.5,4.0)(12.5,-2.5)(0,-2.5)
};
\addplot[red!8, fill=red!8, draw=none, forget plot] coordinates {
  (12.5,-2.5)(12.5,4.0)(45,4.0)(45,-2.5)(12.5,-2.5)
};

\node[font=\scriptsize, text=green!60!black, align=center]
  at (axis cs:5.5, 3.3) {Quant helpful};
\node[font=\scriptsize, text=red!60!black, align=center]
  at (axis cs:30, 3.3) {Quant harmful};

\node[font=\scriptsize, anchor=south west, text=black]
  at (axis cs:8.0, 2.57) {v0.3 (IR=7.7)};
\node[font=\scriptsize, anchor=north, text=violet!80]
  at (axis cs:7.69, 2.14) {v0.4 (IR=7.69)};
\node[font=\scriptsize, anchor=north, text=black]
  at (axis cs:35.1, -1.16) {v0.2 (IR=35.1)};

\addplot[gray!70, thick, dashed, forget plot] coordinates {(12.5,-2.5)(12.5,4.0)};

\end{axis}
\end{tikzpicture}